\documentclass[times, 10pt,twocolumn]{article}
\usepackage{times}
\usepackage{epsfig}
\usepackage{float}
\usepackage{times}
\usepackage{fullpage}
\usepackage{graphicx}
\usepackage{amsmath}
\usepackage{amsfonts}
\usepackage{amssymb}
\usepackage{amstext}
\usepackage{xspace}
\usepackage{theorem}
\usepackage{multicol}
\usepackage{latexsym}
\usepackage[numbers,sort]{natbib}
\usepackage[backref,pageanchor=true,plainpages=false, pdfpagelabels, bookmarks,bookmarksnumbered,
]{hyperref}
\usepackage{subfigure}
\usepackage{supertab}
\usepackage{longtable}

\begin{document}

\title{On Ranking Senators By Their Votes}

\author{Mugizi Rwebangira\\
Howard University\\ Systems and Computer Science \\ 2300 6th Street Washington, DC 20059\\ rweba@cs.cmu.edu\\
}

\maketitle
\thispagestyle{empty}

\begin{abstract}

The problem of ranking a set of objects given some measure of similarity is one of
the most basic in machine learning. Recently \cite{agarwal06} proposed a method
based on techniques in semi-supervised learning utilizing the graph Laplacian. In this
work we consider a novel application of this technique to ranking binary choice data and
apply it specifically to ranking US Senators by their ideology.
\end{abstract}

\section{Introduction}

Ranking is one of the most fundamental problems in machine learning. Over the years, many algorithms
have been proposed many of them relying on classical techniques such as linear regression. In order to
apply regression techniques typically we need have access to the features of the objects in order to
learn the regression equation. We can then run the regression equation on the test data and output our
ranking. \\

Recently there has been more interest in the setting where we only have access to the similarity among objects
and we want to output a ranking. The natural interpretation of this ranking is that we want similar objects to be
ranked close to each other and dissimilar objects to be ranked far apart from each other. This set of similarities
can be visualized as a graph, where the nodes are the objects we want to rank and the weight on the edges expresses
the similarity of the examples connected by that edge. \\

The ``graph learning'' setting has proven very popular in machine learning and there have been a slew of papers
exploring this paradigm \cite{zhu03,zhu05b,belkin06,agarwal06}. In a recent paper Agarwal \cite{agarwal06}
proposed applying the graph learning technique to the problem of ranking. In this paper we propose to apply
this technique to ranking binary choice data, with the specific application of ranking US Senators by their ideology. \\

\section{Background}
\subsection{Ranking Legislators}
The problem of ranking politicians by ideology has been well studied in the political science literature. Currently the most
popular such ranking system is VoteView developed by political scientist Keith Poole and hosted on his web site VoteView.com.
Voteview works by projecting the set of legislators into a low dimensional Euclidean space and then iteratively searching for cutting planes
that optimally divide the legislators into sets that agree with each other \cite{poole08}. \\

In particular we note that this optimization procedure is not guaranteed to converge to a global minimum. By contrast the method
that we present involves only basic linear algebra and is guaranteed to give a global minimum. In addition it is very easy to
implement and has a lower running time. \\

\subsection{The Combinatorial Graph Laplacian}

In our solution we will a matrix known as the \emph{combinatorial graph Laplacian}
(or simply the \emph{Laplacian} of a graph) because of several suggestive analogies with the
classical Laplacian differential operator widely used in physics (among other places). The
Laplacian has many interesting properties of which we will exploit only a few in this work.
For more details refer to the thesis of Zhu\cite{zhu05b}. \\

For an undirected graph $G$, the Laplacian $\Delta$ is defined as the matrix

\[\Delta = D - A \] where $A$ is the adjacency matrix of $G$ and $D$ is the diagonal matrix which has the degree of
each vertex on the diagonal (i.e the degree matrix which is equivalently the sum of very row).

The concept can be straightforwardly generalized to a weighted graph by defining

\[\Delta = D - W \] where $W_{i,j}$ is the weight between nodes $i$ and $j$ and $D_{i,i}$ is the sum of row $i$
of matrix $W$.

\section{Ranking}

Our goal is to develop a method for ranking legislators by ideology based on their votes.
We can use a simple model for this situation: \textbf{Each senator has a \emph{ideology score} $f$ between
$-1$ and $+1$ that represents their political leanings.} For example we can let $+1$ represent
 an extremely liberal senator and $-1$ an extremely conservative one. Our task is to compute these scores
 based on their votes. \\

To derive an algorithm we make a simple and obvious assumption: \textbf{Senators
that have a similar voting record should have similar ideological scores.} That is if two senators
voting records are very similar we assume their political views are also very similar. \\

We propose the following as our objective function which we will seek to minimize.

\[ \mathcal{E}(f) = \sum_i^n \sum_j^n W_{i,j} (f_i-f_j)^2 \]

where $W_{i,j}$ is the similarity between senators $i$ and $j$ and $n$ is the number of senators. \\

To obtain a ranking we have to find the values of $f_i$ which minimize this objective function. \\

\textbf{Lemma 1
}
\[ \mathcal{E}(f) = \sum_i^n \sum_j^n W_{i,j} (f_i-f_j)^2 = 2 f^T\Delta f \]

where $\Delta$ is the graph Laplacian and $f$ is the vector of all the $f_i$'s. \\

\textbf{Proof:}

\[ \mathcal{E}(f) = \sum_i^n \sum_j^n W_{i,j} (f_i-f_j)^2  \]

\[ = \sum_i^n \sum_j^n W_{i,j} (f_i^2-2f_if_j+f_j^2) \]

\[ = \sum_i^n \sum_j^n W_{i,j}f_i^2-2W_{i,j}f_if_j+W_{i,j}f_j^2 \]

\[ = \sum_i^n \sum_j^n W_{i,j}f_i^2-2\sum_i^n \sum_j^nW_{i,j}f_if_j+\sum_i^n \sum_j^nW_{i,j}f_j^2 \]

Now

\[\sum_i^n \sum_j^n W_{i,j}f_i^2 = \sum_i^n  (W_{i,1} + W_{i,2} + \cdots)f_i^2 = f^TDf\]

Where $D$ is a diagonal matrix and $D_{i,i}$ is the sum of row $i$ of matrix $W$. \\

Similarly

\[\sum_i^n \sum_j^n W_{i,j}f_j^2  = f^TDf\]

Here we use the assumption that matrix $W$ is symmetric. \\

Lastly

\[ \sum_i^n \sum_j^n W_{i,j}f_if_j = f^TWf\]

Putting it all together

\[ = \sum_i^n \sum_j^n W_{i,j}f_i^2-2\sum_i^n \sum_j^nW_{i,j}f_if_j+\sum_i^n \sum_j^nW_{i,j}f_j^2 \]

\[ = f^TDf -2f^TWf + f^TDf = 2f^T(D-W)f = 2f^T\Delta f   \]

\textbf{QED} \\

In order to get non-trivial results we will have to specify the values of at least two of the $f_i$.
(Otherwise the minimum can be obtained by just setting all the $f_i$ to zero which is not very useful).
It is convenient to set one $f_i$ to $+1$ and another $f_i$ to $-1$, corresponding to the most ideologically
pure legislators. \\

Once we have done this we can split up the vector $f$ into $f_L$ and $f_U$ corresponding to the labeled $f_i$ and the unlabeled $f_i$ respectively.
Likewise we can rearrange and split up $\Delta$ into $\Delta_{UU}$ $\Delta_{UL}$ $\Delta_{LU}$ $\Delta_{LL}$ as can be seen in the following
table: \\

\begin{tabular}{|c|c|c|}
\hline
&L&U \\ \hline
L&$\Delta_{LL}$&$\Delta_{LU}$ \\ \hline
U&$\Delta_{UL}$&$\Delta_{UU}$ \\ \hline
\end{tabular}
\vskip .25in

Our task now is to minimize $2f^T\Delta f$ which is the same as minimizing $f^T\Delta f$\\

\textbf{Lemma 2
}
The $f_U$ that minimizes  $f^T\Delta f$ is equal to $-\Delta_{UU}^{-1}\Delta_{UL}f_L$ \\

\textbf{Proof:}

\[ f^T\Delta f = [f_L^T f_U^T]   \left( \begin{array}{cc}
\Delta_{LL}&\Delta_{LU} \\
\Delta_{UL}&\Delta_{UU} \\
\end{array} \right)  \left[ \begin{array}{c}
f_{L} \\
f_{U} \\
\end{array} \right] \]

\[ = f_L^T\Delta_{LL}f_{L} + f_L^T\Delta_{LU}f_{U}+ f_U^T\Delta_{UL}f_{L}+ f_U^T\Delta_{UU}f_{U} \]

Differentiating by $f_U$ (since $f_L$ is a constant) and setting to $0$ we get:

\[ 2\Delta_{UL}f_{L} + 2\Delta_{UU}f_{U} = 0\]

Rearranging we get

\[ f_U = -\Delta_{UU}^{-1}\Delta_{UL}f_L \]

\textbf{QED} \\

\section{Algorithm}
To summarize, our algorithm is as follows:

\begin{enumerate}

\item First we compute the similarities between all pairs of examples. This will give us a similarity matrix $W$.

\item We then compute the graph Laplacian $\Delta = D - W$

\item We specify at least two labeled $f_i$ to get $f_L$.

\item We compute $f_U = -\Delta_{UU}^{-1}\Delta_{UL}f_L$ to obtain the final ranking.

\end{enumerate}

Now we just need to specify how we are going to compute the similarities and how we specify
the labeled $f_i$.

\subsection{Computing Similarities}
A legislator can essentially only do 3 things on any particular vote,

\begin{enumerate}

\item Vote ``YES''.

\item Vote ``NO''.

\item Fail to register any vote (e.g. absent, abstaining etc).

\end{enumerate}

We encode the behavior of each legislator as a vector of integers in $\{-1,0,+1\}$ in the obvious way and
define the ``distance'' $D$ between two legislators as the Hamming difference of their respective vectors.
We then define the weight $W_{i,j}$ as $\frac{1}{D+1}$ and thus obtain the weight matrix $W$.

\subsubsection{Example}
Suppose Senator Rightwinger has voted (NO, YES, ABSTAIN) on 3 bills while Senator Leftwinger has
voted (YES,NO,ABSTAIN) on the same set of bills. Then we encode their votes as the vectors (-1,+1,0) and
(+1,-1,0). The Hamming distance of the two vectors is 2 and hence we will assign the similarity between
the two senators the value of $\frac{1}{2+1} = \frac{1}{3}$.

\subsection{Selecting the labeled $f_i$}

In essence we have to specify at least two $f_i$ to which we can confidently assign a label. For our purposes
it makes the most sense to specify the most extreme examples.
There are broadly two ways of doing this
\begin{enumerate}

\item Use domain knowledge of the political arena - e.g. look at the most extreme legislators in
 other rankings by advocacy groups and other parties, look at the political ideology of the legislator's home district and
 other external evidence.

\item Purely internal knowledge from the dataset - For example pick the two legislators with the highest political
difference (lowest similarity).

\end{enumerate}

In practice we find our algorithm is robust to any reasonable choice (i.e the rankings will not change drastically based on the method).

\section{Experimental Results}

We obtained data on roll call votes for the 2007-2008 session of the US Senate from the web site of Keith T. Poole \cite{poole08}.
We removed the votes on which there was more than 95\% agreement as those
were most likely ceremonial votes. \\

\subsection{Using Domain Knowledge}

In this experiment we picked Senator Russell Feingold of Wisconsin and Senator Thomas Coburn of Oklahoma as our liberal
and conservative exemplars. Both of these senators have a strong reputation for exemplifying the liberal and conservative
wings of their respective political parties in the Senate. As per our methodology we fixed $f_i=1$ for Senator Feingold
and $f_i=-1$ for Senator Coburn ran our algorithm and obtained the following results: \\

\begin{table}
\caption{Ranking using the Domain Knowledge Method}
\begin{tabular}{|c|l|c|}
\hline
Rank&Name&Party \\ \hline
1   &   FEINGOLD    & D\\ \hline
2   &   SANDERS     & D\\ \hline
3   &   LEAHY       & D\\ \hline
4   &   DURBIN      & D\\ \hline
5   &   HARKIN      & D\\ \hline
6   &   WYDEN       & D\\ \hline
7   &   BROWN       & D\\ \hline
8   &   WHITEHOUSE  & D\\ \hline
9   &   CARDIN      & D\\ \hline
10  &   MENENDEZ    & D\\ \hline
11  &   KERRY       & D\\ \hline
12  &   CANTWELL    & D\\ \hline
13  &   KOHL        & D\\ \hline
14  &   LAUTENBERG  & D\\ \hline
15  &   KLOBUCHAR   & D\\ \hline
16  &   AKAKA       & D\\ \hline
17  &   MURRAY      & D\\ \hline
18  &   SCHUMER     & D\\ \hline
19  &   REED        & D\\ \hline
20  &   BOXER       & D\\ \hline
21  &   BINGAMAN    & D\\ \hline
22  &   LEVIN       & D\\ \hline
23  &   STABENOW    & D\\ \hline
24  &   REID        & D\\ \hline
25  &   CASEY       & D\\ \hline
26  &   MIKULSKI    & D\\ \hline
27  &   FEINSTEIN   & D\\ \hline
28  &   NELSON      & D\\ \hline
29  &   WEBB        & D\\ \hline
30  &   SALAZAR     & D\\ \hline
31  &   TESTER      & D\\ \hline
32  &   INOUYE      & D\\ \hline
33  &   ROCKEFELLER & D\\ \hline
34  &   KENNEDY     & D\\ \hline
35  &   CONRAD      & D\\ \hline
36  &   DODD        & D\\ \hline
37  &   DORGAN      & D\\ \hline
38  &   CARPER      & D\\ \hline
39  &   BAUCUS      & D\\ \hline
40  &   BIDEN       & D\\ \hline
41  &   MCCASKILL   & D\\ \hline
42  &   LINCOLN     & D\\ \hline
43  &   BYRD        & D\\ \hline
44  &   CLINTON     & D\\ \hline
45  &   LIEBERMAN   & D\\ \hline
46  &   PRYOR       & D\\ \hline
47  &   BAYH        & D\\ \hline
48  &   OBAMA       & D\\ \hline
49  &   LANDRIEU    & D\\ \hline
50  &   NELSON      & D\\ \hline
51  &   JOHNSON     & D\\ \hline
\end{tabular}
\label{table1}
\end{table}

\begin{table}
\caption{Ranking Using the Domain Knowledge Method}
\begin{tabular}{|c|l|c|}
\hline
Rank&Name&Party \\ \hline
52  &   SNOWE       & R\\ \hline
53  &   COLLINS     & R\\ \hline
54  &   SPECTER     & R\\ \hline
55  &   SMITH       & R\\ \hline
56  &   COLEMAN     & R\\ \hline
57  &   WICKER      & R\\ \hline
58  &   VOINOVICH   & R\\ \hline
59  &   THOMAS      & R\\ \hline
60  &   STEVENS     & R\\ \hline
61  &   LUGAR       & R\\ \hline
62  &   MURKOWSKI   & R\\ \hline
63  &   DOMENICI    & R\\ \hline
64  &   WARNER      & R\\ \hline
65  &   HAGEL       & R\\ \hline
66  &   MCCAIN      & R\\ \hline
67  &   LOTT        & R\\ \hline
68  &   COCHRAN     & R\\ \hline
69  &   BENNETT     & R\\ \hline
70  &   HATCH       & R\\ \hline
71  &   BOND        & R\\ \hline
72  &   MARTINEZ    & R\\ \hline
73  &   ROBERTS     & R\\ \hline
74  &   ALEXANDER   & R\\ \hline
75  &   GRASSLEY    & R\\ \hline
76  &   HUTCHISON   & R\\ \hline
77  &   DOLE        & R\\ \hline
78  &   SUNUNU      & R\\ \hline
79  &   BROWNBACK   & R\\ \hline
80  &   CORKER      & R\\ \hline
81  &   CRAIG       & R\\ \hline
82  &   SHELBY      & R\\ \hline
83  &   CRAPO       & R\\ \hline
84  &   BARASSO     & R\\ \hline
85  &   GREGG       & R\\ \hline
86  &   MCCONNELL   & R\\ \hline
87  &   THUNE       & R\\ \hline
88  &   ISAKSON     & R\\ \hline
89  &   CHAMBLISS   & R\\ \hline
90  &   GRAHAM      & R\\ \hline
91  &   VITTER      & R\\ \hline
92  &   CORNYN      & R\\ \hline
93  &   SESSIONS    & R\\ \hline
94  &   BUNNING     & R\\ \hline
95  &   KYL         & R\\ \hline
96  &   ENZI        & R\\ \hline
97  &   BURR        & R\\ \hline
98  &   ALLARD      & R\\ \hline
99  &   ENSIGN      & R\\ \hline
100 &   INHOFE      & R\\ \hline
101 &   DEMINT      & R\\ \hline
102 &   COBURN      & R\\ \hline

\end{tabular}
\label{table2}
\end{table}

\begin{table}
\caption{Ranking Using the Internal Knowledge Method}
\begin{tabular}{|c|l|c|}
\hline
Rank&Name&Party \\ \hline
1   &   MENENDEZ    & D\\ \hline
2   &   LAUTENBERG     & D\\ \hline
3   &   SCHUMER       & D\\ \hline
4   &   DURBIN      & D\\ \hline
5   &   CANTWELL      & D\\ \hline
6   &   MURRAY       & D\\ \hline
7   &   CARDIN       & D\\ \hline
8   &   BROWN  & D\\ \hline
9   &   WHITEHOUSE      & D\\ \hline
10  &   BOXER    & D\\ \hline
11  &   REED       & D\\ \hline
12  &   KERRY    & D\\ \hline
13  &   HARKIN        & D\\ \hline
14  &   SANDERS  & D\\ \hline
15  &   LEAHY   & D\\ \hline
16  &   AKAKA       & D\\ \hline
17  &   BINGAMAN      & D\\ \hline
18  &   LEVIN     & D\\ \hline
19  &   STABENOW        & D\\ \hline
20  &   WYDEN       & D\\ \hline
21  &   KOHL    & D\\ \hline
22  &   FEINSTEIN       & D\\ \hline
23  &   KLOBUCHAR    & D\\ \hline
24  &   MIKULSKI        & D\\ \hline
25  &   REID       & D\\ \hline
26  &   CASEY    & D\\ \hline
27  &   NELSON   & D\\ \hline
28  &   FEINGOLD      & D\\ \hline
29  &   SALAZAR        & D\\ \hline
30  &   WEBB     & D\\ \hline
31  &   KENNEDY      & D\\ \hline
32  &   ROCKEFELLER      & D\\ \hline
33  &   INOUYE & D\\ \hline
34  &   CONRAD     & D\\ \hline
35  &   CARPER      & D\\ \hline
36  &   DORGAN        & D\\ \hline
37  &   TESTER      & D\\ \hline
38  &   BAUCUS      & D\\ \hline
39  &   LINCOLN      & D\\ \hline
40  &   DODD       & D\\ \hline
41  &   BIDEN   & D\\ \hline
42  &   BYRD     & D\\ \hline
43  &   LIEBERMAN        & D\\ \hline
44  &   CLINTON     & D\\ \hline
45  &   PRYOR   & D\\ \hline
46  &   MCCASKILL       & D\\ \hline
47  &   LANDRIEU        & D\\ \hline
48  &   BAYH       & D\\ \hline
49  &   OBAMA    & D\\ \hline
50  &   NELSON      & D\\ \hline
51  &   JOHNSON     & D\\ \hline
\end{tabular}
\label{table3}
\end{table}

\begin{table}
\caption{Ranking Using the Internal Knowledge Method}
\begin{tabular}{|c|l|c|}
\hline
Rank&Name&Party \\ \hline
52  &   SNOWE       & R\\ \hline
53  &   COLLINS     & R\\ \hline
54  &   SPECTER     & R\\ \hline
55  &   SMITH       & R\\ \hline
56  &   COLEMAN     & R\\ \hline
57  &   WICKER      & R\\ \hline
58  &   THOMAS   & R\\ \hline
59  &   VOINOVICH      & R\\ \hline
60  &   STEVENS     & R\\ \hline
61  &   MURKOWSKI       & R\\ \hline
62  &   LUGAR   & R\\ \hline
63  &   MCCAIN    & R\\ \hline
64  &   DOMENICI      & R\\ \hline
65  &   WARNER       & R\\ \hline
66  &   HAGEL      & R\\ \hline
67  &   LOTT        & R\\ \hline
68  &   COCHRAN     & R\\ \hline
69  &   HATCH     & R\\ \hline
70  &   ROBERT       & R\\ \hline
71  &   BENNETT        & R\\ \hline
72  &   MARTINEZ    & R\\ \hline
73  &   ALEXANDER     & R\\ \hline
74  &   BOND   & R\\ \hline
75  &   GRASSLEY    & R\\ \hline
76  &   BROWNBACK   & R\\ \hline
77  &   HUTCHISON        & R\\ \hline
78  &   CORKER      & R\\ \hline
79  &   DOLE   & R\\ \hline
80  &   SUNUNU      & R\\ \hline
81  &   BARASSO       & R\\ \hline
82  &   CRAIG      & R\\ \hline
83  &   SHELBY       & R\\ \hline
84  &   CRAPO     & R\\ \hline
85  &   THUNE       & R\\ \hline
86  &   ISAKSON   & R\\ \hline
87  &   GREGG       & R\\ \hline
88  &   MCCONNELL     & R\\ \hline
89  &   CHAMBLISS   & R\\ \hline
90  &   GRAHAM      & R\\ \hline
91  &   SESSIONS      & R\\ \hline
92  &   VITTER      & R\\ \hline
93  &   CORNYN    & R\\ \hline
94  &   BUNNING     & R\\ \hline
95  &   ENZI         & R\\ \hline
96  &   BURR        & R\\ \hline
97  &   ALLARD        & R\\ \hline
98  &   KYL      & R\\ \hline
99  &   ENSIGN      & R\\ \hline
100 &   INHOFE      & R\\ \hline
101 &   COBURN      & R\\ \hline
102 &   DEMINT      & R\\ \hline
\end{tabular}
\label{table4}
\end{table}

\subsection{Using Internal Knowledge}

We did another experiment where we picked as our two exemplars the senators who were the \textbf{least}
 similar in terms of their voting records. In this case our algorithm ended up picking Senator Robert
 Menendez of New Jersey ande Senator Jim DeMint of South Carolina. This is interesting because while
 Senator DeMint has a reputation in the Senate as a staunch conservative, Senator Menendez does not have as
 high a public profile. This analysis suggests his voting record may be more partisian than his low profile reputation
 suggests.  As per our methodology we fixed $f_i=1$ for Senator Menendez
and $f_i=-1$ for Senator DeMint and ran our algorithm and obtained the results shown on the previous page. \\

\section{Discussion}

First we note that the rankings produced are very reasonable and correlate well with rankings produced by interest groups
and political commentators. The advantage of an data driven method of course is that it does not require human expertise.
Secondly the ``domain knowledge'' and ``internal knowledge'' methods produce very similar results. This suggests a
certain degree of robustness. The ``internal evidence'' method appears preferable as it does
not require any choice of parameters.

\subsection{Conclusions}
We have presented a fast method for ranking legislators based on their votes. The method gives reasonable results,
is easy to implement and apparently more straightforward than competing methods such as that of Poole. In addition
the similarity matrix is an intuitive concept and suggests some applications in the area of visualizing the legislature.

\subsection{Future Work}
One interesting idea is to explore the idea of using different similarity functions. In this work we used the plain
vanilla Hamming distance, about the simplest things that we could use. It is possible that a more sophisticated domain specific
similarity function might produce qualitatively different result (e.g. diffusion kernel or Rank Similarity). Another idea
is to further explore any significant qualitative differences with other ranking algorithms to establish the respective
advantages and disadvantages of the various methods.

\nocite{*}
\bibliographystyle{latex8}
\bibliography{icdm09}

\begin{thebibliography}{1}\setlength{\itemsep}{-1ex}\small

\bibitem{agarwal06}
S.~Agarwal.
\newblock Ranking on graph data.
\newblock In {\em Proceedings of the 23rd International Conference on Machine
  Learning}, 2006.

\bibitem{belkin06}
M.~Belkin, P.~Niyogi, and V.~Sindhwani.
\newblock Manifold regularization: A geometric framework for learning from
  labeled and unlabeled examples.
\newblock {\em Journal of Machine Learning Research}, 7:2399--2434, 2006.

\bibitem{chapelle06}
O.~Chapelle, B.~Sch{\"o}lkopf, and A.~Zien, editors.
\newblock {\em Semi-Supervised Learning}.
\newblock MIT Press, Cambridge, MA, 2006.

\bibitem{poole08}
K.~T. Poole.
\newblock Voteview.com.
\newblock http://www.voteview.com, 2008.

\bibitem{zhu05a}
X.~Zhu.
\newblock Semi-supervised learning literature survey.
\newblock Technical Report 1530, Computer Sciences, University of
  Wisconsin-Madison, 2005.
\newblock http://www.cs.wisc.edu/$\sim$jerryzhu/pub/ssl\_survey.pdf.

\bibitem{zhu05b}
X.~Zhu.
\newblock Semi-supervised learning with graphs.
\newblock 2005.
\newblock Doctoral Dissertation.

\bibitem{zhu03}
X.~Zhu, Z.~Ghahramani, and J.~Lafferty.
\newblock Semi-supervised learning using {G}aussian fields and harmonic
  functions.
\newblock In {\em Proceedings of the 20th International Conference on Machine
  Learning}, pages 912---919, 2003.

\end{thebibliography}

\end{document}